%% file: root.tex
\documentclass[letterpaper, 10 pt, conference]{ieeeconf}  

\IEEEoverridecommandlockouts                              

\usepackage{multicol}
\usepackage[bookmarks=false]{hyperref}
\usepackage[pdftex]{graphicx}
\usepackage{amsmath}
\usepackage{amssymb}
\usepackage{amsthm}
\usepackage{bm}
\usepackage{mathrsfs}
\usepackage{float}
\usepackage{stfloats}
\usepackage{array}
\usepackage{tabu}
\usepackage{boldline}
\usepackage{multirow}
\usepackage{hhline}
\usepackage{color, colortbl, xcolor}
\usepackage[flushleft]{threeparttable}

\newcolumntype{?}{!{\vrule width 1.5pt}}

\usepackage[lofdepth,lotdepth]{subfig}
\usepackage{lipsum}
\usepackage[symbol]{footmisc}

\title{\LARGE \bf
Learning the Latent Space of Robot Dynamics \\for Cutting Interaction Inference
}

\author{Sahand Rezaei-Shoshtari$^{\: 1 \; 2}$, David Meger$^{\: 1 \; 2}$, Inna Sharf$^{\: 3}$
\thanks{$^{1}$School of Computer Science, McGill University, Montreal, Canada $^{2}$Mila Quebec AI Institute $^{3}$Department of Mechanical Engineering, McGill University, Montreal, Canada. Correspondence to: Sahand Rezaei-Shoshtari 
        {\tt\small sahand.rezaei-shoshtari@mail.mcgill.ca}}%
}

\begin{document}

\maketitle
\thispagestyle{empty}
\pagestyle{empty}

\begin{abstract}
Utilization of latent space to capture a lower-dimensional representation of  a complex dynamics model is explored in this work. The targeted application is of a robotic manipulator executing a complex environment interaction task, in particular, cutting a wooden object. We train two flavours of Variational Autoencoders---standard and Vector-Quantised---to learn the latent space which is then used to infer certain properties of the cutting operation, such as whether the robot is cutting or not, as well as, material and geometry of the object being cut. The two VAE models are evaluated with reconstruction, prediction and a combined reconstruction/prediction decoders. The results demonstrate the expressiveness of the latent space for robotic interaction inference and the competitive prediction performance against recurrent neural networks.

\end{abstract}

\section{Introduction}

The research presented in this paper is motivated by the need to increase intelligence and automation in timber-harvesting applications. These are operations which involve a heavy-duty multi-degree-of-freedom arm, mounted on a mobile base and equipped with sophisticated end-effectors, which are employed to saw and manipulate wood. The interaction dynamics that occur in these operations, in particular in the wood-cutting tasks,  are complex and present significant challenges for model learning approaches.  An accurate representation of the dynamics of such interactions can enable  more intelligent control of the arm and motion planning for the whole machine. 



This paper aims to tackle some of the challenges related to model learning of robotic arms carrying out  a cutting operation. To this end, we use the Kinova Jaco arm equipped with a specialized end-effector as a small-scale proxy for a timber-harvesting machine. In particular, we hypothesize that a latent space, an unobserved space inferred from the raw sensory data of a robot, can provide valuable, yet concise information on its interactions, even in a complex cutting task. We further extend this notion to be able to predict its future states to render the model learning framework useful for model-based reinforcement learning and control.

\begin{figure}[t!]
    \centering
    \includegraphics[width=0.45\textwidth]{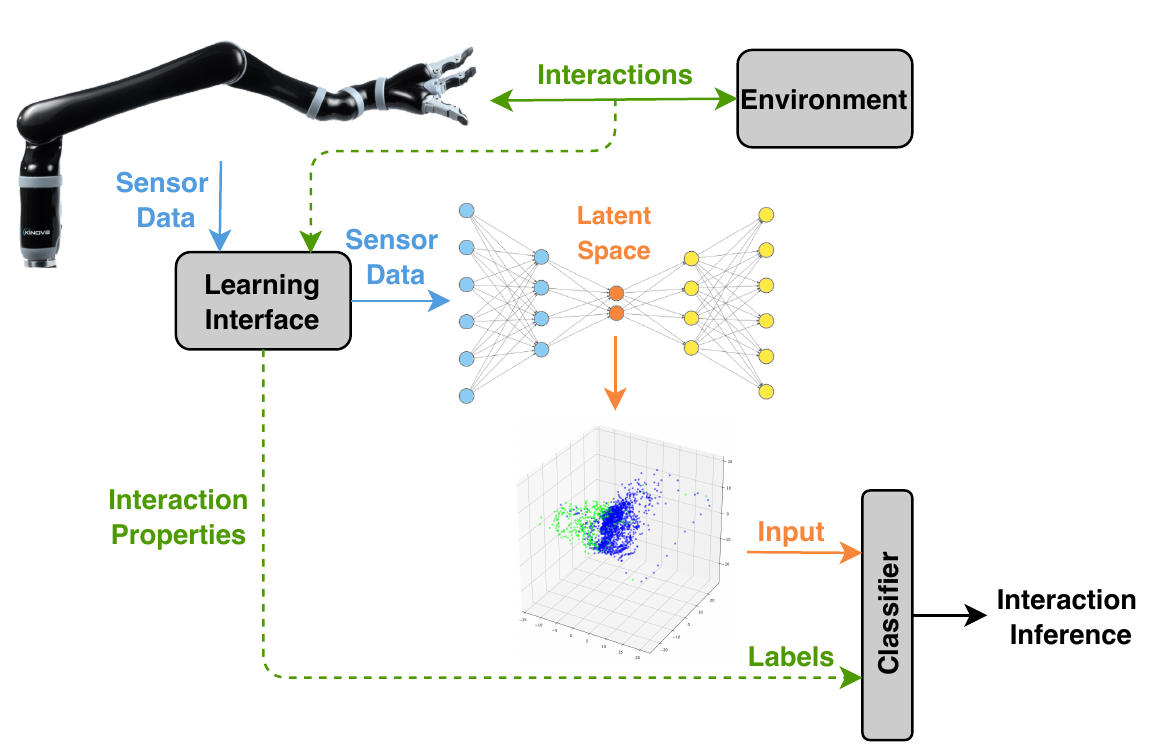}
    \caption[Overview of the latent space learning and classification framework.]{Overview of the latent space learning and classification framework. First, a latent space is learned from the robot sensor data using Variational Autoencoders and then samples from the latent space are used as the inputs for training a classifier for inferring robot interactions.}
    \label{fig:latent_learning_diagram}
\end{figure}

The inherent strength of latent variable models in capturing representations has made them particularly useful for learning complex dynamics models of systems with unobserved states and interactions. Time-lagged Variational Autoencoders \cite{hernandez2018variational} include a time-lag element into the architecture in order to compress complex nonlinear systems in a single embedding with high fidelity to the underlying dynamics. The time-lag enables the model to act as a standard forward model capable of predicting the future. Neural Relational Inference \cite{kipf2018neural} learns to infer the interactions of a system while simultaneously learning its dynamics from observational data; the latent code represents the interactions of the system in the form of a graph neural network. 

Recently, a significant shift towards the use of latent dynamics models has emerged in model-based reinforcement learning \cite{hafner2019learning, watter2015embed, kaiser2019model, gelada2019deepmdp, zhang2019solar}. Learning the control policies in the latent dynamics space, rather than the original high-dimensional state space, often results in better generalization \cite{hafner2019learning}. In the same context, deep generative models have been utilized to learn and predict physical interactions of a robotic arm from raw video data \cite{finn2016unsupervised}. 

Besides neural networks, Gaussian processes have also received considerable attention in the context of latent variable models \cite{lawrence2004gaussian}. Latent force models  \cite{alvarez2013linear} have been proposed as a hybrid approach between Gaussian processes and ordinary differential equations; these physically driven kernels are defined based on the underlying interactions of the system and are thus able to better capture them. Stochastic control policies for these models have also been derived for controlling second-order systems with unknown input signals \cite{sarkka2018gaussian}.

Closely related to the approach taken in this work, the latent space of dynamics has been substantiated as a powerful tool for capturing physical interactions of a robot, particularly in cutting tasks. DeepMPC \cite{lenz2015deepmpc} proposes an MPC framework for robotic manipulators using recurrent neural networks for learning unobserved properties of food cutting dynamics. Similarly, an MPC formulation for velocity controlled robots has been derived based on recurrent neural networks incorporating force/torque measurements of an external sensor \cite{mitsioni2019data}. Furthermore, auxiliary tasks, such as thickness prediction of the cutting material, have been shown to help learning an embedding space of the dynamics model from images in the context of vegetable slicing \cite{sharma2019learning}.

Differently from previous approaches explored in the context of robotic cutting, our framework uses Variational Autoencoders instead of recurrent models and, furthermore, it only uses proprioceptive measurements of joint states of the robot. With an introduction of a time-lagged element into the decoder alongside the standard reconstruction, we demonstrate the expressiveness of latent space for interaction inference, as well as forward prediction.

We evaluate our proposed method on a series of actual robotic cutting tasks on a variety of wood specimen most commonly found in timber-harvesting operations. To the best of our knowledge, this is the first attempt at learning representations in the context of robotic wood cutting with a saw. In addition, we have released an extensive experimental dataset generated over the course of this work\footnote{The dataset and its details are available at \href{https://github.com/McGill-AML/aml\_robot\_cutting\_dataset}{https://github.com/McGill-AML/aml\_robot\_cutting\_dataset}}.

We continue with Section II, which provides a background on Variational Autoencoders; section III describes the proposed framework for learning interaction-rich robot dynamics models, while Section IV evaluates this method by presenting the details of the experiments and the obtained results. Finally, section V summarizes our work.

\section{Background}
\subsection{Variational Autoencoders}
Generative latent variable models aim to learn the probability distribution of a dataset by simultaneously learning the distribution of its unobserved lower-dimensional representations. Assuming $x \in \mathcal{X}$ and $z \in \mathcal{Z}$ as the full state and latent space, respectively, samples are then generated from a learned joint distribution of $p(\mathbf{x}, \mathbf{z}) = p(\mathbf{z}) p(\mathbf{x}|\mathbf{z})$, where vectors $\mathbf{z}$ and $\mathbf{x}$ are respectively sampled from a prior distribution $p_\theta(\mathbf{z})$ and a conditional distribution $p_\theta(\mathbf{x}|\mathbf{z})$. Imposing a prior distribution on the latent variables ensures the smoothness of the distribution. 

Applying Bayes' rule gives the true posterior distribution of the latent variables as $p_\theta(\mathbf{z}|\mathbf{x}) = p_\theta(\mathbf{x}|\mathbf{z}) p_\theta(\mathbf{z}) / p_\theta(\mathbf{x})$, where the denominator can be calculated by integrating the joint distribution $p(\mathbf{x}, \mathbf{z})$  over the latent space $\mathcal{Z}$ to obtain the marginal likelihood $p_\theta(\mathbf{x})$ as:
\begin{equation}
    p_\theta(\mathbf{x}) = \int p_\theta(\mathbf{z}) p_\theta(\mathbf{x}|\mathbf{z}) d\mathbf{z}.
    \label{eq:vae}
\end{equation}
The underlying task is to maximize likelihood of $\mathcal{X}$, however, the integration and the true posterior $p_\theta(\mathbf{z}|\mathbf{x})$ are intractable. Variational Autoencoders (VAE) \cite{kingma2013auto} tackle this issue by introducing a recognition model $q_\phi(\mathbf{z}|\mathbf{x})$ as an approximation to the true and intractable posterior. The recognition model functions as a probabilistic encoder by mapping the inputs into the latent variables, while the generative model $p_\theta(\mathbf{x}|\mathbf{z})$ acts as a probabilistic decoder, by reconstructing the original inputs conditioned on the latent codes. 

Rearranging and rewriting the marginal likelihood yields Equation (\ref{eq:vae_loss}) as the loss function in VAEs:
\begin{align}
    \label{eq:vae_loss}
    \mathcal{L} (\theta, \phi, \mathbf{x}^{(i)}) =& -D_{KL}\big(q_\phi(\mathbf{z}|\mathbf{x}^{(i)}) || p_\theta(\mathbf{z)} \big) \\ \nonumber
    &+ \mathbb{E}_{q_\theta (\mathbf{z}|\mathbf{x}^{(i)})} \big[ \text{log} \: p_\theta(\mathbf{x}^{(i)}| \mathbf{z}) \big],
\end{align}
where the first term is the KL-divergence between the approximate and true posterior and acts as a regularization term for measuring the amount of information lost by approximating the true posterior. The second term is the reconstruction loss for measuring the expectation of the likelihood of the reconstructed data with respect to the distribution over the latent variables provided by the encoder \cite{doersch2016tutorial}.

In VAEs, both the distribution over the latent variables $p_\theta(\mathbf{z})$ and the approximate posterior $q_\theta(\mathbf{z}|\mathbf{x})$ are assumed to be a multivariate Gaussian. This assumption on the approximate posterior allows for applying the ``reparametrization trick'' by implementing the posterior as $\text{log} \: q_\theta(\mathbf{z}|\mathbf{x}^{(i)}) = \text{log} \: \mathcal{N}(\mathbf{z}; \mathbf{\mu}^{(i)}, \mathbf{\sigma}^{2(i)} \mathbf{I})$ \cite{kingma2013auto}.

\subsection{Vector-Quantised Variational Autoencoders}
One of the major shortcomings of VAEs is that the prior on the latent space is considered to be a Gaussian; this assumption is maintained throughout the learning process by imposing the KL-divergence term in the loss function as defined in Equation (\ref{eq:vae_loss}). As a result, their performance can decrease when this assumption is not satisfied. On the other hand, Vector-Quantised Variational Autoencoders (VQ-VAE) \cite{van2017neural} do not impose the said assumption and learn a discrete latent representation while simultaneously learning the prior.

In VQ-VAEs, the prior and posterior distributions are categorical and the samples drawn from the latent representation index an embedding space $e \in \mathbb{R}^{K \times D}$ where $K$ is the size of the discrete latent space and $D$ is the dimensionality of each latent embedding vector. The input $\mathbf{x}$ is passed to the model and the encoder creates the output $\mathbf{z}_e(\mathbf{x})$; the discrete latent variables $\mathbf{z}$ are in turn calculated using a nearest neighbour look-up in the embedding space $e$. 

In general, discrete latent embeddings can be a potent substitute in cases where the latent variables are discrete by nature. Additionally, from the empirical standpoint, it has been demonstrated that VQ-VAEs do not suffer from the ``posterior collapse'' which is often observed in VAEs. Posterior collapse happens when the latent variables are ignored by the decoder when it is powerful enough to reconstruct the original inputs, regardless of the latent embeddings \cite{van2017neural}.

\section{Methodology}
\label{sec:methods}
Traditionally, for modeling and control of robotic arms with active interactions, the forces have been identified and included in the rigid body dynamics model \cite{featherstone2016dynamics}, stated as:
\begin{equation}
    \mathbf{H}(\bm{q}) \ddot{\bm{q}} + \mathbf{C}(\bm{q}, \bm{\dot{q}}) \bm{\dot{q}} + \bm{\tau}_g(\bm{q}) = \bm{\tau} + \bm{\tau}_{\varepsilon} + \mathbf{J}^\intercal(\bm{q}) \bm{F},
    \label{eq:dynamics}
\end{equation}
where $\bm{q} \in \mathbb{R}^{N}$ is the vector of joint positions, $\mathbf{H}(\bm q) \in \mathbb{R}^{N \times N}$ is the inertia matrix, $\mathbf{C}(\bm q, \bm{\dot{q}})\bm{\dot{q}} \in \mathbb{R}^{N}$ represents Coriolis and centrifugal forces, $\bm{\tau}_g(\bm q) \in \mathbb{R}^{N}$ denotes gravitational effects, $\bm \tau \in \mathbb{R}^{N}$ is the vector of joint torques, $\bm{\tau}_\varepsilon$ is the additional torque due to friction and disturbances, $\mathbf{J}(\bm{q})$ is the Jacobian matrix with respect to the end-effector, and $\bm{F}$ is the interaction force and moment applied to the end-effector.

In the context of cutting interactions, a common approach in contact dynamics is to approximate the interaction force as a spring-damper system of the form \cite{featherstone2016dynamics}: 
\begin{equation}
    \bm{f} = \mathbf{K_p}(\bm{x} - \bm{x_r}) + \mathbf{K_d}(\bm{\dot{x}} - \bm{\dot{x}_r}),
    \label{eq:interaction_force}
\end{equation}
where $\bm{x}$ and $\bm{x_r}$ are respectively the current and reference positions of the blade inside the specimen. The values for $\mathbf{K_p}$, $\mathbf{K_d}$, and $\bm{x_r}$ are dependant on the properties of the specimen (e.g., material and thickness) or configuration of the cutting tool. However, capturing these unobservable parameters of the environment can be challenging. Instead of utilizing System Identification tools \cite{wu2010overview}, we propose to learn a latent space that encodes these unobservable features alongside a  dynamics model in a unified framework for inferring interactions and predicting future states.


\subsection{Latent Space of Robot Dynamics}
\begin{figure}[t!]
    \centering
    \includegraphics[width=0.45\textwidth]{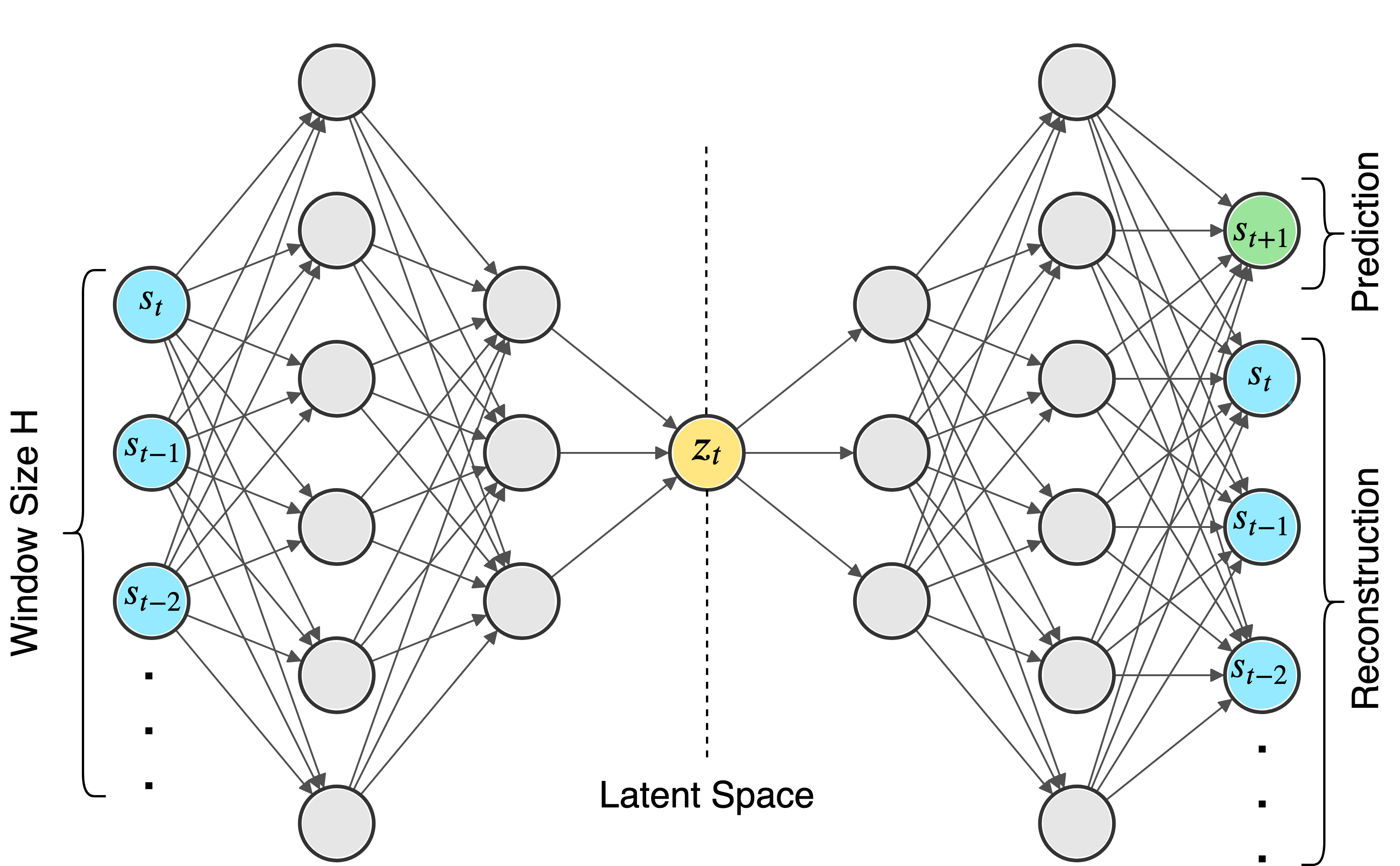}
    \caption{Variational Autoencoder as a dynamics model. The inputs are the the states over a window of size $H$.}
    \label{fig:dynamics_model_encoder}
\end{figure}

We utilize Variational Autoencoders to learn the latent space of cutting dynamics. VAEs are usually trained via reconstruction of input vectors using a reconstructive decoder network. However, in the context of dynamics models, reconstruction of current robot joint states, albeit useful for learning expressive representations, does not allow for prediction of future states, which is essential for model-based reinforcement learning and control. Thus, we propose to modify the decoder to predict the next states alongside the standard reconstruction, as shown in Figure \ref{fig:dynamics_model_encoder}. During training, the loss function is then extended to impose the prediction error in addition to the reconstruction and KL-divergence losses. The prediction loss uses the representation to capture the transition model of the underlying dynamics, while the reconstruction and KL-divergence losses ensure expressiveness of these representations, without which the performance of interaction inference is decreased significantly as demonstrated in Section \ref{sec:evaluation}.

In this work, the input vectors of the dynamics model $\mathbf{x} \in \mathbb{R}^M$ are defined as the manipulator joint states (positions, velocities and torques) aggregated over the last $H$ time-steps. Therefore, for a robot arm with $N$ degrees of freedom, the input is a $\mathbf{x} \in \mathbb{R}^{3 \times H \times N}$ vector. Furthermore, a reconstruction decoder, has $\mathbf{x}' \in \mathbb{R}^{3 \times H \times N}$ output vectors, whereas a prediction decoder outputs $\mathbf{x}' \in \mathbb{R}^{2 \times N}$ vectors, as it predicts only the joint positions and velocities of the next state. 

\subsection{Classification with Deep Generative Models}
The approach taken here is closely related to the methods seen in semi-supervised learning frameworks. Latent-feature discriminative models (M1 models) are commonly used in this context and are based on the idea that a learned model capable of providing embeddings or feature representations of the data, can significantly simplify the classification task by clustering the related observations in a latent space \cite{kingma2014semi}. 

Taking the same approach, we first learn the latent space or latent embeddings of robot dynamics with VAE and VQ-VAE models and consequently learn a classifier for predicting class labels from this latent representation. As discussed in \cite{kingma2014semi}, the lower dimensionality of the latent space as well as the fact that the non-linear transformation of the data allows for its higher moments to be captured by the latent space, improve the classification performance. Figure \ref{fig:latent_learning_diagram} shows an overview of the learning framework; support vector machines (SVM) have been used as the classifier.

\section{Evaluation}
\label{sec:evaluation}
In this section, we evaluate the performance of the framework described earlier for inferring robot interactions and predicting future states in interaction-rich environments.

\subsection{Experimental Setup}
Motivated by the timber-harvesting application, the experiments in this research are carried out on a series of real robotic cutting tasks with different wood specimens. We have collected a Robot Cutting Dataset consisting of approximately 1500 seconds, a total of 180 cuts (15 cuts on 12 unique wood specimens) using Kinova Jaco arm retrofitted with a custom end-effector fixture and saw. The wooden rods have been selected from sets of five materials and five thicknesses (see Table \ref{table:cutting_dataset}). The test bed, Figure \ref{fig:exp_setup}, consists of four components:

\begin{figure}[b!]
    \centering
    \includegraphics[width=0.4\textwidth]{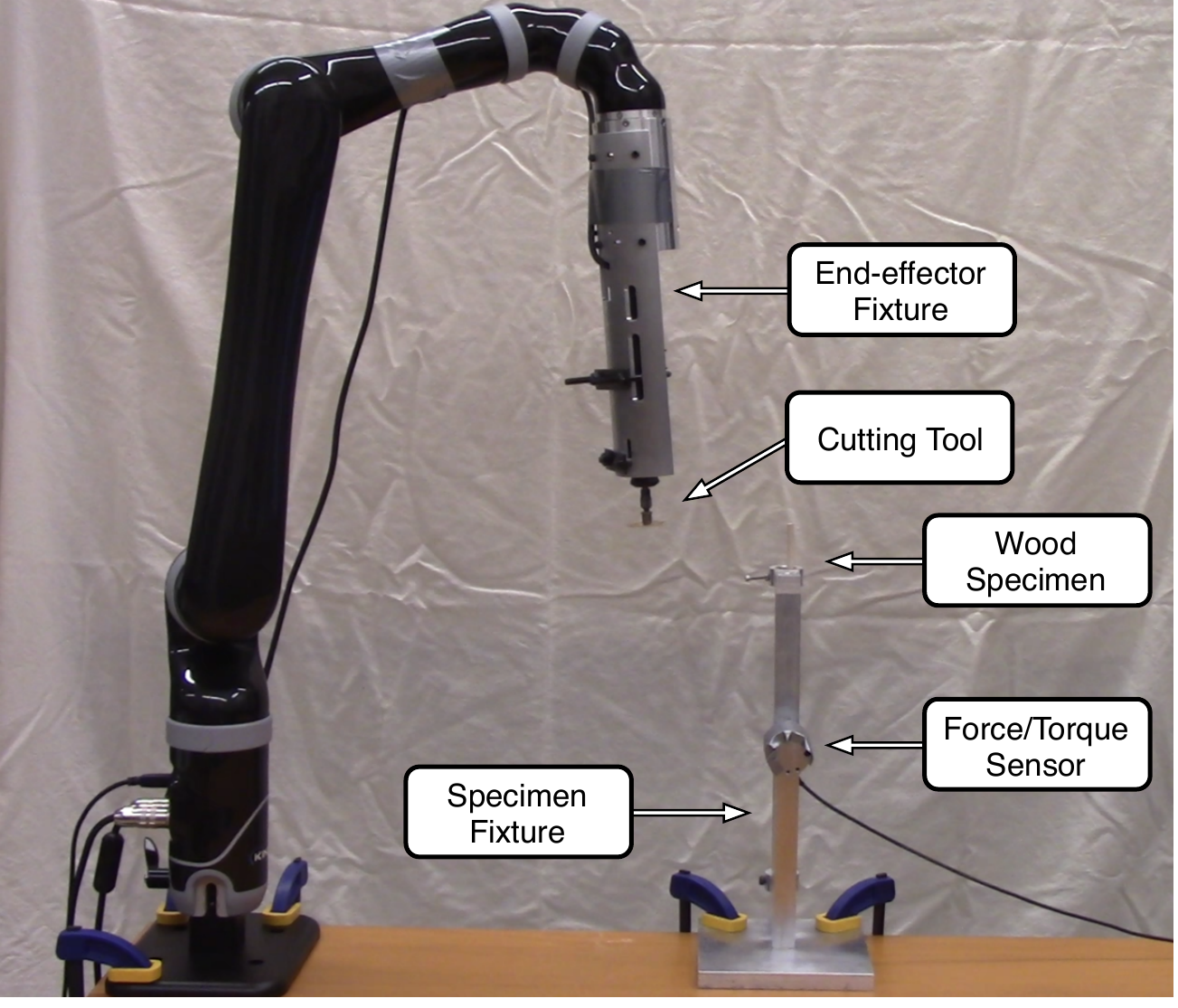}
    \caption{Robot cutting test bed used for data collection. A custom end-effector fixture replaces the original gripper.}
    \label{fig:exp_setup}
\end{figure}

\begin{itemize}
    \item \textbf{Kinova Jaco 2 Arm:} A light-weight six-DOF robotic arm equipped with joint encoders and joint torque sensors for collecting data at 10 Hz.
    
    \item \textbf{Cutting Tool and End-effector Fixture:} Designed as a replacement for the original three-finger gripper, the aluminum end-effector fixture holds the cutting tool. The tool is a Dremel 100 Single Speed Rotary Tool Kit, equipped with a circular saw blade.
    
    \item \textbf{Specimen Fixture:} This component provides a firm support for holding the wood specimen and measuring the cutting force, using an integrated force/torque sensor in the middle of the fixture. The latter is not used in our work because integration of sensors on a timber-harvesting machine is highly challenging.
\end{itemize}

\begin{figure*}[b!]
  \centering
    \begin{tabular}[c]{cc}
        \multirow{2}{*}[110pt]{
        \subfloat[][Interaction inference and classification.]{
            \centering
            \includegraphics[width=0.69\textwidth]{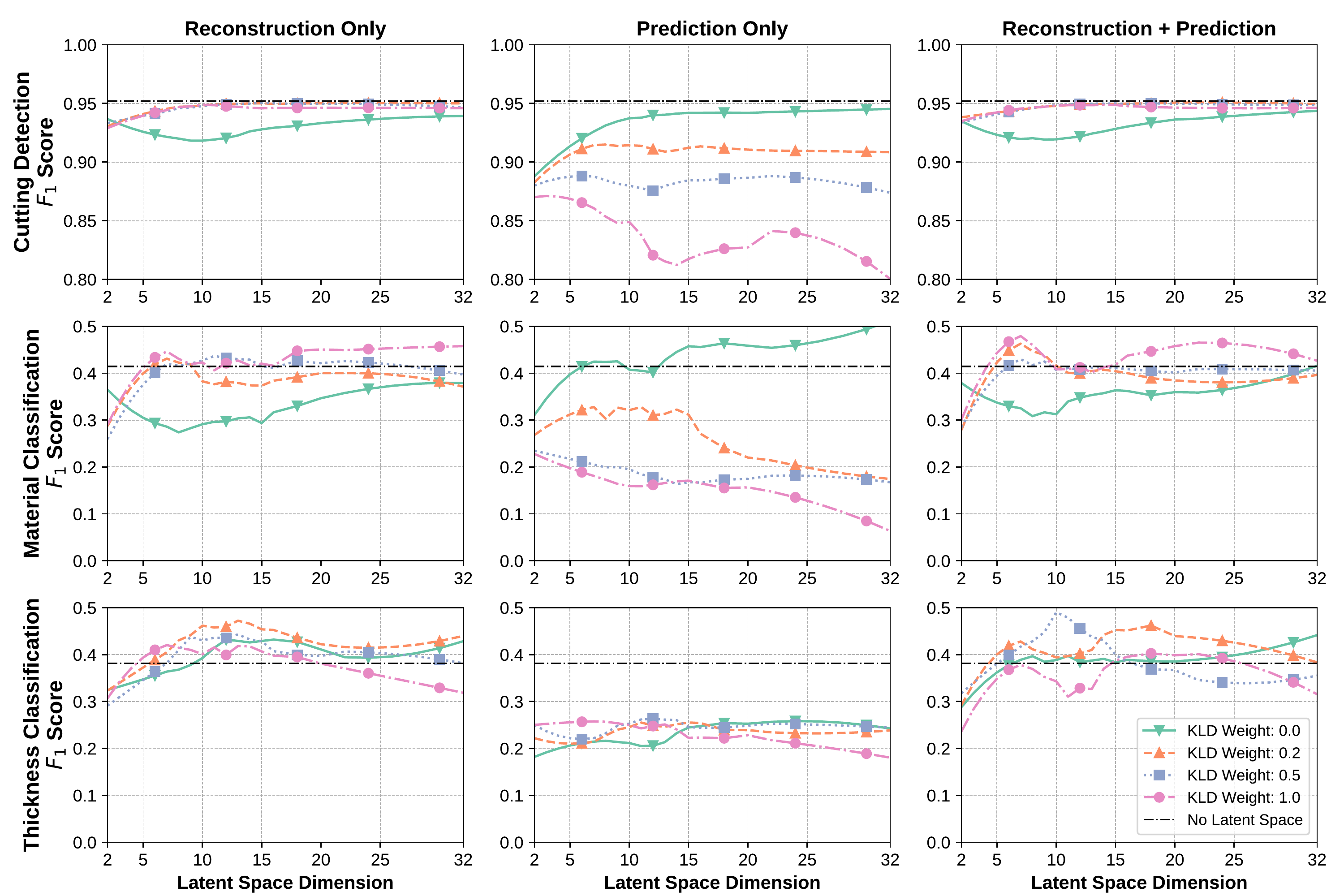}
            \label{fig:vae_classification_results}}}
        & \subfloat[][Forward prediction.]{
            \centering
            \includegraphics[width=0.22\textwidth]{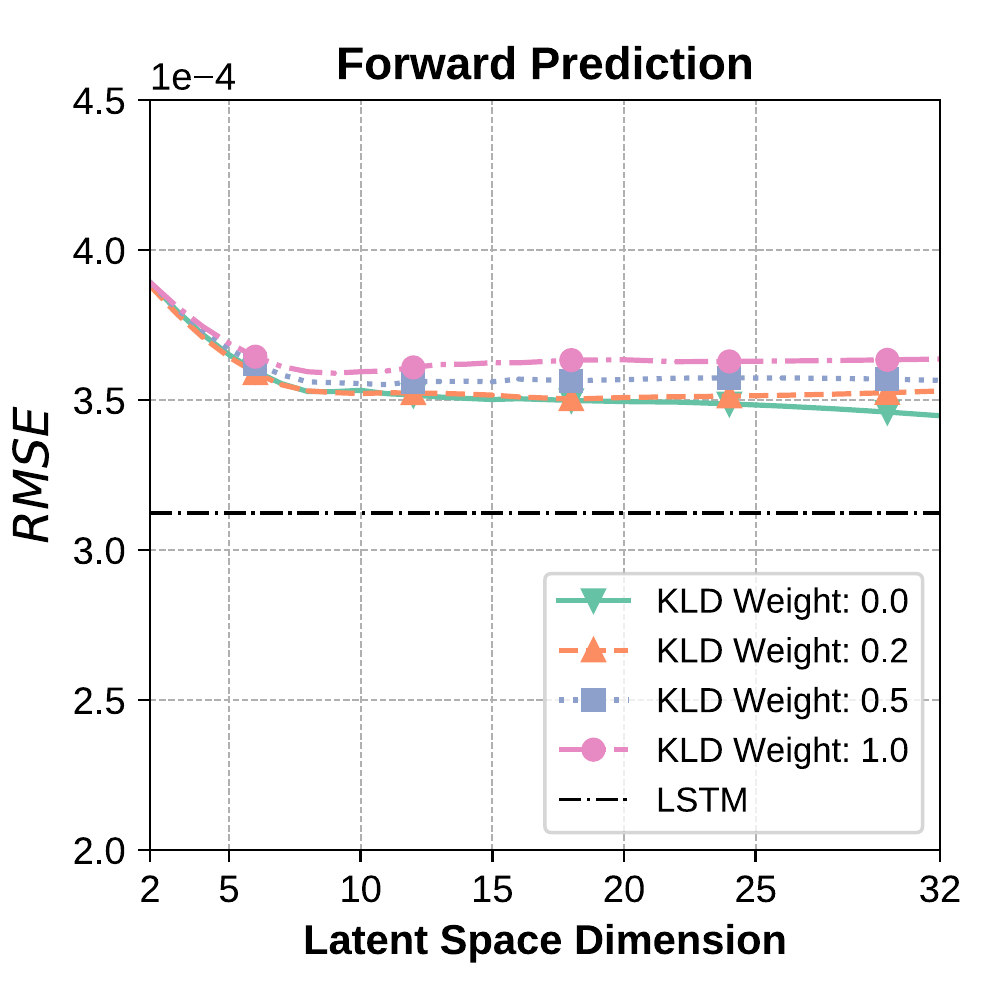}
            \label{fig:vae_prediction_results}} \\
        & \subfloat[][Latent space visualization.]{
            \centering
            \quad\includegraphics[width=0.19\textwidth]{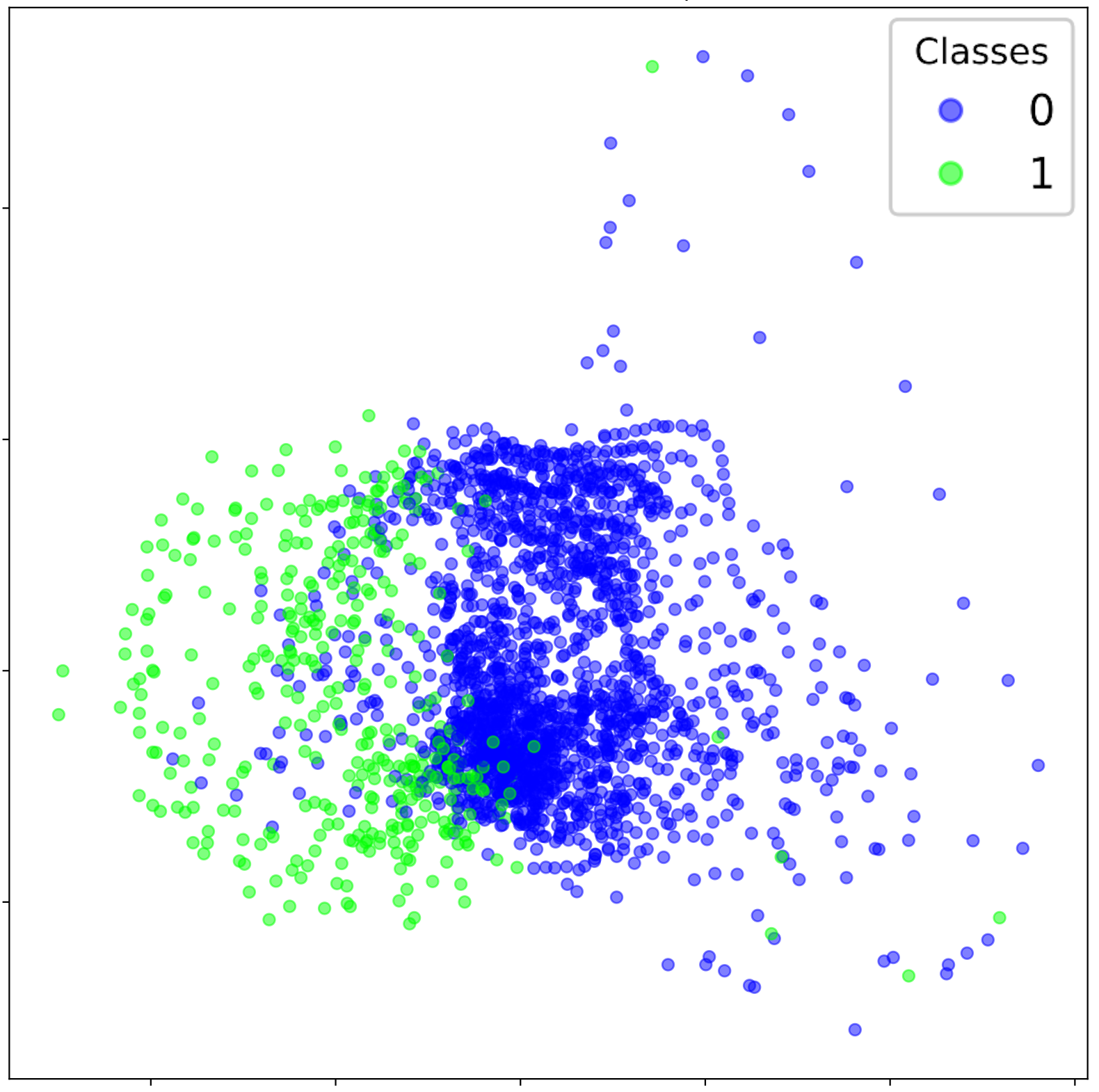}
            \label{fig:vae_viz}}
    \end{tabular}
    \setlength{\abovecaptionskip}{6pt}
    \caption{Performance of VAE models for different dimensions of the latent space, evaluated on the test set. \textbf{(a)} Interaction inference using the latent space of VAE models, separated by the classification task at each row (cutting detection, material classification, and thickness classification) and the setup of the VAE at each column (reconstruction only, prediction only, and both). The baseline method is classification directly on the state space. \textbf{(b)} RMSE of forward prediction of a VAE model. The baseline method is an LSTM network. \textbf{(c)} Visualization of a 4-dimensional latent space of a VAE model. The latent vectors have been projected to 2-dimensional space using PCA; the blue circles indicate free-motion and the green circles represent a cutting-motion.}
\end{figure*}

During each cut, the robot is moved manually from the start position, Figure \ref{fig:exp_setup}, towards the wood specimen in Cartesian velocity control. In the case of stiff and thick specimens, a cut is not possible with a single pass and the tool needs to be retracted for a second try. This multi-modality imposes further challenges for latent space learning.

\begin{table}[b!]
    \centering
    \caption{Details of wood specimens used in the dataset.}
    \input{dataset_table}
    \label{table:cutting_dataset}
\end{table}

The dataset is split into three subsets for training ($80\%$), validation ($10\%$), and test ($10\%$). Furthermore, for labeling the dataset and identifying robot's engagement in a cutting interaction, we have compared the moving mean and standard deviation of joint torques over a window \cite{kay1993fundamentals}: a sudden increase in the variance can be interpreted as a start or end time of a cutting episode. 

The network architectures are multilayer perceptron (MLP) with two hidden layers, each of dimension 40, with standard Rectified Linear Unit (ReLU) \cite{nair2010rectified}. We trained the models using the Adam optimizer \cite{kingma2014adam} with learning rate $5\mathrm{e}\scriptstyle{-}\displaystyle{4}$ and batch size 128. The data has been normalized to have zero mean and standard deviation of 1.

\begin{figure*}[b!]
  \centering
    \begin{tabular}[c]{cc}
    \multirow{2}{*}[125pt]{
        \subfloat[][Interaction inference and classification.]{
        \includegraphics[width=0.67\textwidth]{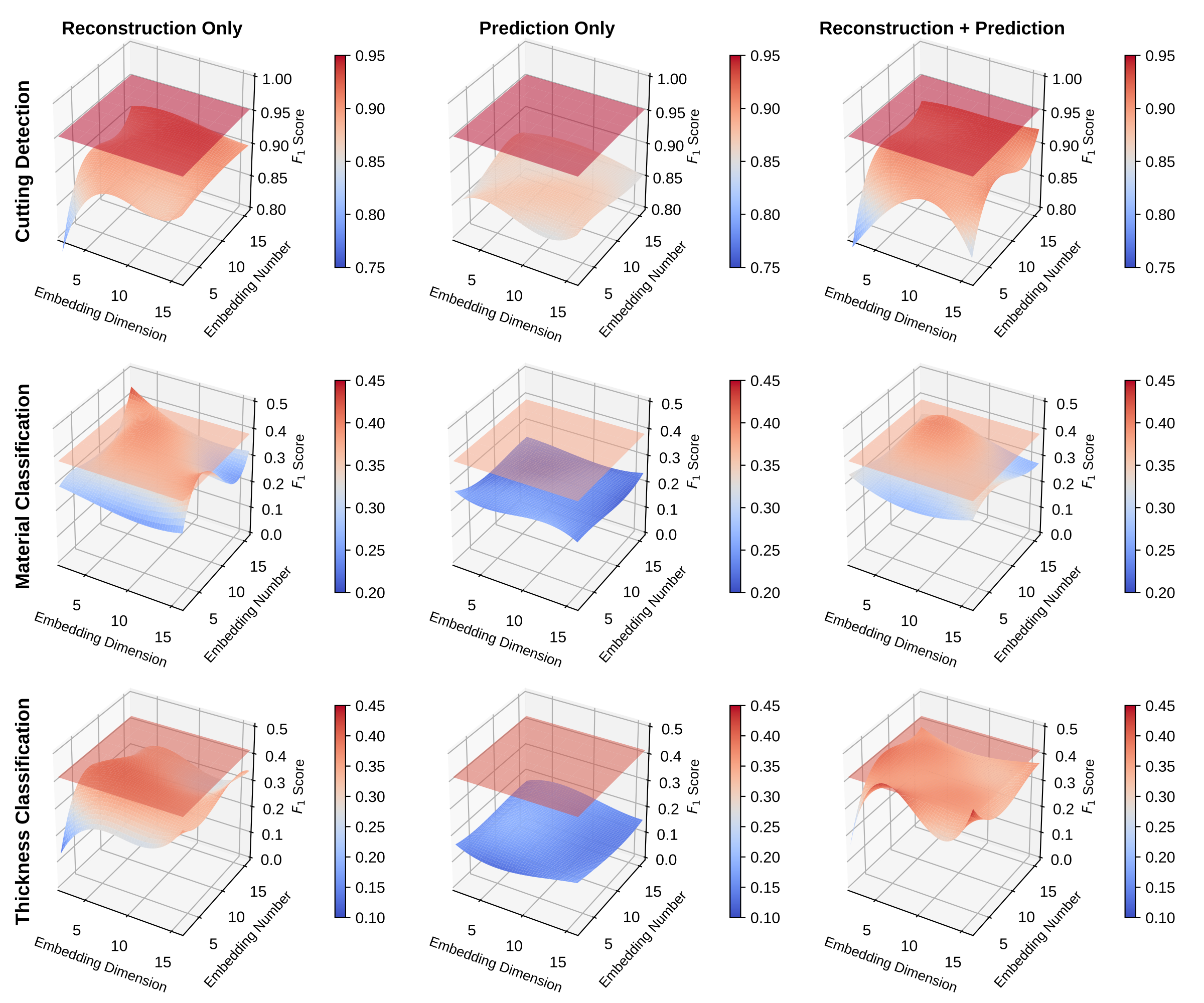}
        \label{fig:vqvae_classification_results}}}
        & \subfloat[][Forward prediction.]{
            \quad\includegraphics[width=0.24\textwidth]{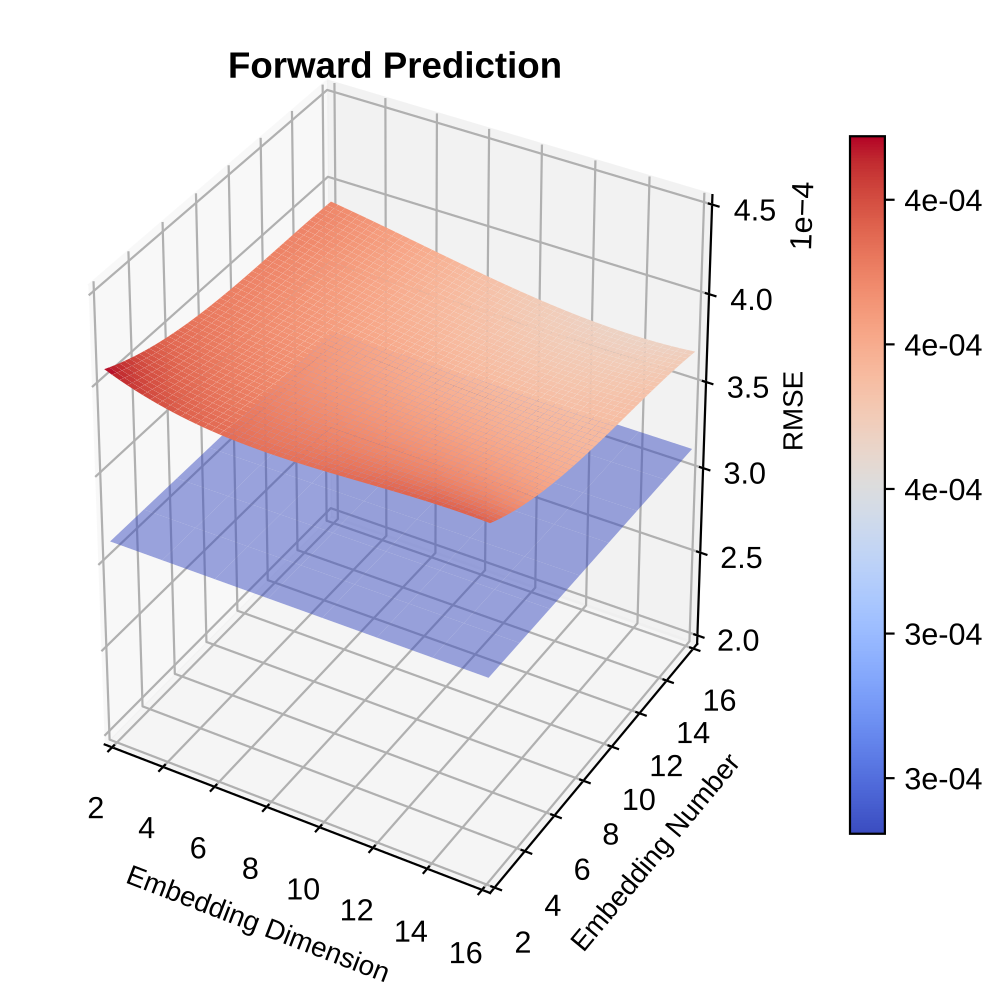}
            \label{fig:vqvae_prediction_results}} \\ \\ \\
        & \subfloat[][Latent space visualization.]{
            \includegraphics[width=0.18\textwidth]{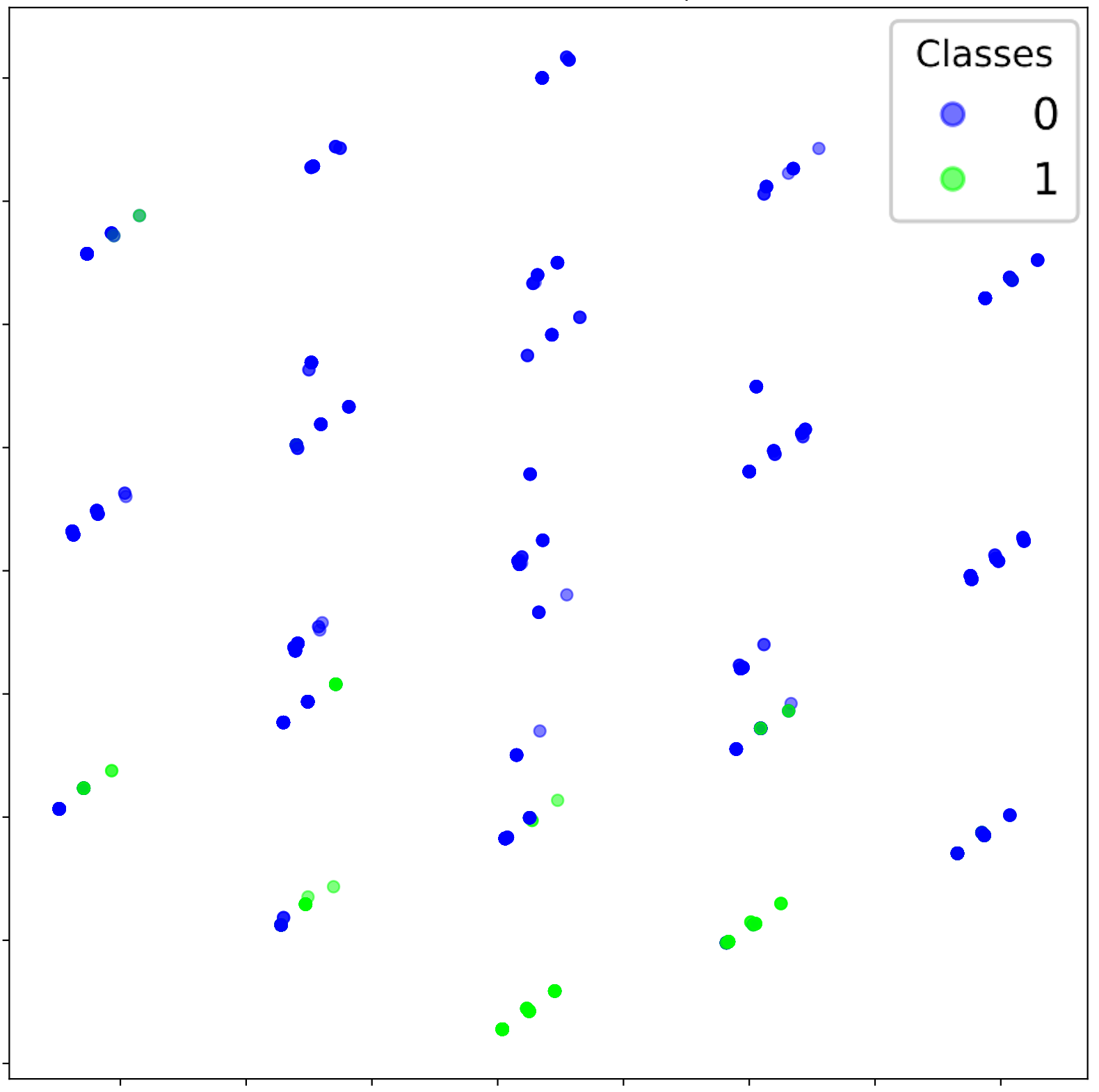}
            \label{fig:vqvae_viz}}
    \end{tabular}
    \setlength{\abovecaptionskip}{22pt}
    \caption{Performance of VQ-VAE models for different dimensions and numbers of the latent embeddings, evaluated on the test set. \textbf{(a)} Interaction inference using the latent space of VQ-VAE models, separated by the classification task at each row (cutting detection, material classification, and thickness classification) and the setup of the VAE at each column (reconstruction only, prediction only, and both). The baseline method is classification directly on the state space. \textbf{(b)} RMSE of forward prediction of a VQ-VAE model. The baseline method is an LSTM network. \textbf{(c)} Visualization of a 4$\times$4-dimensional latent space of a VAE model. The latent vectors have been projected to 2-dimensional space using PCA; the blue circles indicate free-motion and the green circles represent a cutting-motion.}
\end{figure*}

\subsection{Interaction Inference}
\label{sec:interaction_inference}
As an evaluation of the quality and expressiveness of the latent space of robot dynamics, we solve a classification problem for inferring the properties of robot interactions. Namely, we consider three classification tasks: (1) cutting detection for inferring whether the robot is engaged in a cut, (2) material classification, and (3) thickness classification. 

Figure \ref{fig:vae_classification_results} shows the classification performance of SVM models trained on the latent representation of VAE models, as a function of latent space dimension for various values of KL-divergence weight in the loss function. Figure \ref{fig:vqvae_classification_results} presents the classification performance on the latent embeddings learned by VQ-VAEs, as a function of embedding dimension and number. The performance is measured using the standard $F_1 = 2/\big({p^{-1} + r^{-1}} \big)$ score which is the harmonic mean of precision $p$ and recall $r$. In order to fully investigate the importance of reconstruction and prediction decoders in learning expressive latent representations, each classification task has been tested with three implementations, the results for which are organized in respective columns: (1) reconstructing the current joint states, (2) predicting the next joint states, and (3) simultaneous reconstruction and prediction. 

As the interaction inference results for prediction-only models suggest, learning the latent space by solely predicting the next state of the robot does not provide an interpretative representation of the state for robotic interactions. This could be the result of a ``mode collapse'': since the output of the predictive decoder is of much lower dimension compared to that of the reconstructive decoder, the latent space is ignored by the predictive decoder as it can predict accurately without relying on the latent space. Combining the two, as suggested in Section \ref{sec:methods}, ensures both the interpretability of the latent space, and the predictive ability of the decoder, thus making it useful for model-based reinforcement learning and control. Moreover, the classification performance is increased slightly compared to reconstruction-only models, as employing a predictive model helps with capturing the time-dependency of the system in the latent space.

In addition, the outperformance of the proposed framework against the baseline method, classification on the state space, emphasizes the fact that the lower dimensionality of the latent space is able to capture higher moments of data and simplify the classification task. Figures \ref{fig:vae_viz} and \ref{fig:vqvae_viz} present the visualization of the latent space of VAE and VQ-VAE models as an example of the interpretability of the latent space.

Table \ref{table:classification_results} compares the $F_1$ scores obtained on the best latent variable models. The best model in each category is chosen respectively based on the validation loss during training and is thus only a representative of the latent space learning performance. Notwithstanding the effect of KL-divergence weight on the performance of VAE models, they outperform VQ-VAEs in general, despite their simplifying assumptions. The continuous domain of dynamics models can be an explanation for this, as it can be better captured through a continuous latent space.
Further optimization of the KL-divergence in VAEs, can also improve their performance.

\setlength{\textfloatsep}{0pt}
\begin{table}[t!]
    \small
    \centering
    \begin{threeparttable}
    \begin{center}
    \vspace{3pt}
    \caption{$F_1$ scores of the best models selected during validation and evaluated on the test set.}
    \setlength\tabcolsep{3pt}
    \begin{tabular}[c]{l *5{c}}
        \hlineB{2}
        & \multicolumn{2}{c}{\textbf{VAE}} & \multicolumn{2}{c}{\textbf{VQ-VAE}} & \multirow{2}{*}{\textbf{Baseline}$^{*}$} \\
        & Dim. & Score & $\text{Dim.}\times\text{Num.}$ & Score & \\
        \hlineB{2}
        Cutting & 11 & \textbf{.9526} & $2\times10$ & .8979 & .9521 \\ 
        \hline
        Material & 4 & .4222 & $6\times8$ & \textbf{.4623} & .4143\\ 
        \hline
        Thickness & 6 & \textbf{.5107} & $10\times12$ & .3842 & .3817\\ 
        \hlineB{2}
    \end{tabular}
    \label{table:classification_results}
    \begin{tablenotes}
        \item $^{*}$The baseline is classification directly on the state space.
    \end{tablenotes}
    \end{center}
    \end{threeparttable}
\end{table}

Finally, it is worth mentioning that the generally low scores on material and thickness classification tasks are due to the inherent difficulty of inferring these values by relying only on the robot joint states.

\subsection{Prediction}
In order to reliably apply a dynamics model for planning and control, the forward prediction errors need to be substantially low. As an evaluation of the prediction performance of the proposed approach, we compare it with recurrent neural networks, namely LSTMs \cite{hochreiter1997long}. We have used a sequence of robot states with the same length as the window size $H$ as the inputs to a two-layer stacked LSTM network with a hidden dimension of 40 and ReLU activation units. We conducted an active search for hyperparameter values to  tune the network and noticed negligible performance changes.

Figures \ref{fig:vae_prediction_results} and \ref{fig:vqvae_prediction_results} show the root-mean-square error of forward predictions for VAE and VQ-VAE models as a function of the latent space size. In Table \ref{table:prediction_results}, we summarize the RMSE of the best validation models chosen during training, the same models as in the first row of Table \ref{table:classification_results}.

\setlength{\textfloatsep}{-2pt}
\begin{table}[t!]
    \small
    \centering
    \vspace{3pt}
    \begin{center}
    \caption{RMSE of forward prediction of the best models selected during validation and evaluated on the test set.}
    \setlength\tabcolsep{3pt}
    \begin{tabular}[c]{l *5{c}}
        \hlineB{2}
        & \multicolumn{2}{c}{\textbf{VAE}} & \multicolumn{2}{c}{\textbf{VQ-VAE}} & \multirow{2}{*}{\textbf{LSTM}} \\
        & Dim. & RMSE & $\text{Dim.}\times\text{Num.}$ & RMSE & \\
        \hlineB{2}
        Prediction & 11 & $3.54{e}\scriptstyle{-}\displaystyle{4}$ & $2\times10$ & $3.93{e}\scriptstyle{-}\displaystyle{4}$ & $\mathbf{3.12{e}\scriptstyle{-}\displaystyle{4}}$ \\
        \hlineB{2}
    \end{tabular}
    \label{table:prediction_results}
    \end{center}
\end{table}

As the results show, LSTM models marginally outperform VAE and VQ-VAE models. Despite the fact that latent variable models use a simpler network architecture without any recurrent feedback, they can achieve competitive performance for modeling robot dynamics with unobserved interactions, thus showcasing the importance of latent space learning. In addition, the latent space of these models can be directly used for inferring robot interactions as discussed in Section \ref{sec:interaction_inference}; whereas, there is no explicit access to the latent space of recurrent neural networks for such use cases, other than their saved memory. 

\section{Conclusion}
In this paper, we explored the notion of latent space of robot dynamics models, particularly in the case of complex and interaction-rich environments. We demonstrated that by modifying a Variational Autoencoder to predict the future states of a robot, alongside the reconstruction of the current states, an expressive latent space can be learned directly from the raw sensory joint states data over a finite window. To evaluate our approach, we collected a dataset from a series of robotic wood cutting tasks with a retrofitted Kinova Jaco arm. We showed that the learned latent representations can encode the information required for drawing inference on robot's interactions. Furthermore, we showed that the prediction capabilities of the proposed model are competitive with 
recurrent neural networks, while providing more interpretative information through the latent space. Future work may explore recurrent autoencoders and use of predictions in model-predictive control.

\section*{Acknowledgments}
This work was supported by the National Sciences and Engineering Research Council (NSERC) Canadian Robotics Network (NCRN).

\bibliographystyle{IEEEtran.bst}
\bibliography{IEEEabrv.bib,references}

\end{document}

%% file: dataset_table.tex
\small
\begin{center}
\begin{tabular}[c]{c *{5}{c}}
    \hlineB{2}
    \multirow{2}{*}{\textbf{Material}} & \multicolumn{5}{c}{\textbf{Thickness (in)}} \\
    & 3/16 & 1/4 & 5/16 & 3/8 & 7/16 \\
    \hlineB{2}
    {LVL} & $\text{\rlap{$\checkmark$}}\square$ & $\text{\rlap{$\checkmark$}}\square$ & $\text{\rlap{$\checkmark$}}\square$ & $\text{\rlap{$\checkmark$}}\square$ & $\text{\rlap{$\checkmark$}}\square$ \\ 
    \hline
    {Maple} & $\square$ & $\text{\rlap{$\checkmark$}}\square$ & $\square$ & $\text{\rlap{$\checkmark$}}\square$ & $\square$\\ 
    \hline
    {Oak} & $\square$ & $\text{\rlap{$\checkmark$}}\square$ & $\square$ & $\text{\rlap{$\checkmark$}}\square$ & $\square$\\ 
    \hline
    {Birch} & $\square$ & $\text{\rlap{$\checkmark$}}\square$ & $\square$ & $\text{\rlap{$\checkmark$}}\square$ & $\square$\\ 
    \hline
    {Hardwood} & $\square$ & $\square$ & $\square$ & $\text{\rlap{$\checkmark$}}\square$ & $\square$\\ 
    \hlineB{2}
\end{tabular}
\end{center}